\documentclass[runningheads]{llncs}

\usepackage[T1]{fontenc}
\usepackage{graphicx}
\usepackage{booktabs}
\usepackage{amsmath}
\usepackage{array}
\usepackage{float}
\usepackage{xcolor}
\usepackage[hidelinks]{hyperref}
\usepackage{orcidlink}
\usepackage[capitalize]{cleveref}

\renewcommand{\orcidID}[1]{\,\orcidlink{#1}}

\newcommand{\arcone}{ARC-AGI-1}
\newcommand{\passone}{pass@1}
\newcommand{\passtwo}{pass@2}

\begin{document}

\title{Pretraining and Test-Time Training Budgets Matter for ARC: A Controlled Study\\ of VARC and TRM}
\titlerunning{Training Budgets Matter for ARC}

\author{Adrien Deli\`ege\orcidID{0000-0003-3981-6982}\thanks{Corresponding author.} \and Marc Van Droogenbroeck\orcidID{0000-0001-6260-6487}}
\authorrunning{A. Deli\`ege and M. Van Droogenbroeck}
\institute{Montefiore Institute of Electrical Engineering and Computer Science,\\
University of Li\`ege, Li\`ege, Belgium\\
\email{adrien.deliege@uliege.be}}

\maketitle

\begin{abstract}
The Abstraction and Reasoning Corpus (ARC) is increasingly used to evaluate whether neural networks can infer task-specific transformations from few demonstrations. Recent ARC models differ in architecture, but also in broader training and evaluation protocols, including pretraining data, data augmentation, test-time setup, and optimization budgets, making conclusions about architecture benefits debatable. In this work, we study how the performance of the vision-inspired VARC and recursion-based TRM models is affected by pretraining and test-time training (TTT) budgets under a common inductive protocol on \arcone{}. We cross five pretraining budgets with four TTT budgets for each test task. Each budget specifies a task-variant count and total sample presentations, defining the controlled training budgets. We observe that increasing TTT sample presentations improves the standard ARC \passtwo{} metric in all settings for both models. Increasing the TTT task-variant count consistently helps only TRM, whereas VARC often performs better with fewer variants. VARC also remains competitive without pretraining, while TRM benefits much more from it. These results show that pretraining and TTT budgets affect ARC evaluation and motivate reporting at comparable budgets.

\end{abstract}

\section{Introduction}
\label{sec:intro}

\begin{figure}[!htbp]
  \centering
  \includegraphics[width=\textwidth]{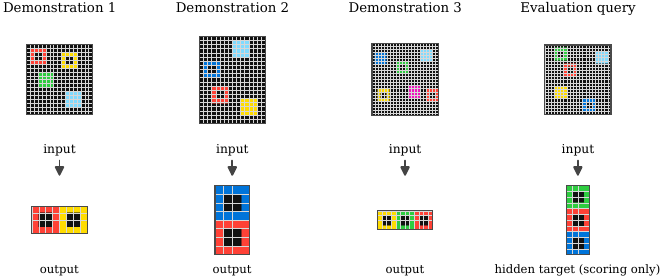}
  \caption{\textbf{Example of an ARC task.} Evaluation task \texttt{a680ac02} provides three demonstration input-output pairs specifying a task-specific transformation. The evaluation query exposes only its input; the target needs to be predicted.}
  \label{fig:arc-anatomy}
\end{figure}

The Abstraction and Reasoning Corpus (ARC) was proposed as a benchmark composed of hundreds of abstract tasks to measure how reasoning capabilities acquired from training tasks generalize on unseen test tasks from few data samples~\cite{chollet2019measure}. Each ARC task contains a small number of training (or ``demonstration'') input-output pairs and a few test inputs, all consisting of grids of integers, usually visualized as colored grids (see \Cref{fig:arc-anatomy}). A solver (human or model) must infer the task-specific transformation rule that maps the training inputs to their outputs and then use that rule to produce the correct test output grids. This setting has made ARC challenging for many years because it requires the ability to efficiently learn new rules at test time without human intervention.

The expectation of ARC creators was that models would be trained on training tasks and then evaluated on test tasks, as in conventional machine learning but with each task acting as a data unit. Pretrained large language models have also been evaluated on ARC~\cite{arcprize2026leaderboard}, despite ARC not being a language-oriented problem~\cite{xu2024llmsarc}. However, unknown training data sources make it difficult to assess whether their performance is driven by on-the-fly reasoning or familiarity with similar tasks in internet-scale datasets. Together with the prohibitive cost of retraining such models, this motivates us to focus on models that can be trained from scratch, in the original ARC spirit, under controlled experimental conditions and that still achieve useful ARC performance. This leads us to two model families with reusable code: VARC~\cite{hu2025arcvisionproblem} and TRM~\cite{jolicoeurmartineau2025morerecursivereasoningtiny}, from which later work has borrowed training and test-time training (TTT) procedures and sometimes architectures. Their pretraining and TTT budgets differ and are often overlooked when only final performance is compared. These budget choices are legitimate, but they make it difficult to tell whether a performance gap is due to architecture or training resources. We therefore investigate a simple question:

\begin{quote}
\emph{How do VARC and TRM perform on ARC under various equivalent pretraining and test-time training budgets?}
\end{quote}

What exactly are these budgets? First, we use the term \emph{protocol} for the general complete training and evaluation procedure, including the data source, pretraining and TTT setup, task-variant construction, prediction aggregation and voting procedure. Then, within a protocol, each optimization stage (pretraining or TTT) has a \emph{budget configuration}: a \emph{task-variant count}, i.e., how many task augmentations are created to diversify the training, and a \emph{sample-presentation budget}, i.e., the total number of demonstrations presented during optimization. We study these pretraining and TTT budget variables while leaving the rest of model-specific elements of the training protocols intact. Our \textbf{findings} are direct:
\begin{itemize}
    \item \textbf{TTT sample-presentation budget matters consistently.} Raising it improves performance in every case for both models.
    \item \textbf{TTT task-variant count matters for TRM.} More variants help TRM broadly, but do not reliably improve VARC's top-ranked predictions.
    \item \textbf{VARC relies less on pretraining.} TTT without pretraining is especially damaging to TRM, whereas VARC remains competitive.
\end{itemize}

As a result, we encourage future ARC papers to benchmark models under fixed pretraining and TTT budgets and to report them explicitly, using our code at \url{https://github.com/AdrienDeliege/arc-training-budgets/tree/main}.

Importantly, our goal is not to decide whether VARC or TRM is intrinsically better. We use the strongest VARC and TRM variants reported in their corresponding papers: VARC has 18M parameters, whereas TRM has 7M, and this capacity difference alone can affect performance. We instead use them as two model families for studying budget sensitivity. Nor do we aim to introduce a new model. Comparable budget measurements are useful without a new state-of-the-art architecture: ARC experiments can consume tens or hundreds of GPU-hours (see~\cref{sec:compute}), so our results can help researchers with limited academic compute choose a budget configuration before scaling a model.

\section{Background and Related Work}
\label{sec:background}

\subsubsection{ARC evaluation.}
ARC tasks consist of grids of integers with values from 0 to 9, usually visualized as colored grids with color 0 in black serving as ``background''. Each task usually provides 3 to 5 training input-output demonstration pairs and 1 (sometimes 2 or 3) test inputs. In ARC evaluations, solvers (humans or models) typically submit up to 2 candidate outputs per test input. If all the test inputs of a task are correctly predicted cell-wise, the task is considered solved at \passtwo{}. ARC contains 400 training tasks to optionally pretrain a model, and 400 test tasks. The proportion of test tasks solved at \passtwo{} is the final performance. The metric \passone{} is also sometimes reported to imagine the case where just one prediction is allowed, as well as an \emph{oracle} metric: for each test example, oracle is correct if the correct grid appears among all generated candidate outputs.

\subsubsection{Visual ARC models.}
Some ARC works have shown that representation choices matter: object-based and spatial encodings can change the difficulty faced by large language models~\cite{li2025vitarc,xu2024llmsarc}. Interestingly, VARC frames ARC as visual image-to-image translation~\cite{hu2025arcvisionproblem}. Grids are rendered onto a fixed canvas with special background and border tokens, then processed by a vision-inspired model such as U-Net~\cite{ronneberger2015unet} or ViT~\cite{dosovitskiy2021image}. We will use VARC-ViT-18M in our paper and refer to it just as ``VARC''. VARC uses ARC training tasks for pretraining as well as hundreds of thousands of external generated examples using the RE-ARC library~\cite{hodel2023rearc}, followed by independent TTT on each test task. During TTT, augmented versions of the tasks are used to adapt task-specific parameters and model weights on their own training samples, before running inference on their test inputs, aggregating predictions and voting for the most promising ones. Although RE-ARC greatly improves performance, we exclude it here to keep the comparison restricted to ARC training data. Building on top of VARC, Embed-TTT~\cite{deliege2026implicitruleinductiontesttime} shows that, at test time, finetuning only a task-specific token already solves a significant fraction of the tasks, while further finetuning the whole backbone in a two-stage setting yield moderate yet consistent improvements, while showing that the structure of the task token space captures rich semantic information about the transforms underlying each task. In addition, LoopViT~\cite{shu2026loopvit} adopts the principle of looped transformers to reuse variants of the same transformer block, achieving slightly better results with fewer parameters. The training and TTT protocols are the same as VARC. Their results show that relatively shallow visual architectures can solve a large subset of \arcone{}, without establishing which reasoning mechanisms individual tasks require. Finally, Nat-ARC~\cite{Ding2026NatARC} leverages an actual vision model (not a vision-inspired model that processes tokenized integer grids like previous works), pretrained in a self-supervised way on ImageNet~\cite{DengImagenet} and fined-tuned per-task during TTT on ARC tasks, to boost abstract reasoning capabilities of a VARC-like model.

\subsubsection{Recursive ARC models.}
TRM uses a compact recursive model over tokenized grids~\cite{jolicoeurmartineau2025morerecursivereasoningtiny}, simplifying a recent line of recurrent models triggered by HRM~\cite{wang2025hrm}. It maintains latent states and recursively refines its prediction, aiming to trade architectural size for iterative computation. Leveraging the compounding effect of two nested loops, TRM is very compact: the version we use has only 7M parameters, yet unrolling its full computational graph reveals hundreds of forward layer passes. Building on top of TRM, FPRM~\cite{movahedi2026fixedpointreasonersstableadaptive} provides theoretical and experimental tools to facilitate model training with its recurrent core and less handcrafted stopping criteria for the adaptive computation time that recurrence imposes. These recurrent approaches are appealing for ARC because many tasks look algorithmic or iterative. They do not use RE-ARC data. Instead, they perform task augmentations during the training process, and encode the test tasks concurrently, making the overall approach transductive (i.e. the training process can in principle already adapt to the future test tasks) rather than inductive (i.e. where test tasks are never seen before being tackled, like in VARC). For our controlled comparison, we use the inductive approach, and independent per-task TTT so that each test task is processed individually. 
Other ARC models use object-centric search or minimum-description-length objectives rather than the neural TTT protocols studied here~\cite{ferre2023objectcentric,liao2025arcagiwithoutpretraining}. We view these methods as complementary; our experimental scope is neural models whose reported performance depends on pretraining and per-task TTT.

\subsubsection{Test-time training (TTT).}
Test-time training adapts a model on data available at evaluation time before making a prediction~\cite{sun2020test}. ARC is a natural setting for TTT because each test task explicitly includes demonstrations. However, ARC TTT is not merely adapting to an input distribution; it attempts to infer a new transformation from a few paired examples. The TTT setup is therefore part of the evaluation protocol. A model trained from scratch on one evaluation task with hundreds of task variants faces a different adaptation burden from a pretrained model fine-tuned on a dozen variants. Although pretraining budgets are commonly examined, TTT budgets receive much less attention~\cite{akyurek2025the}. Our experiments make both budget types explicit.

\section{Budget Definitions and Experimental Setup}
\label{sec:protocol}

\subsection{Budget Measure and Reference Configurations}

\subsubsection{Budget configuration: task variants and sample-presentation budget.} We distinguish the general training \emph{protocol} from the controlled \emph{budget configurations}. The protocol comprises the complete training and evaluation procedure: data sources, pretraining or scratch initialization, TTT setup, task-variant construction, optimization parameters, prediction views, voting scheme, etc. Both VARC and TRM have their own protocols, which we keep mostly intact except for two quantities that we investigate: the task-variant count $V$ and the sample-presentation budget $B$. A \emph{task variant} is a geometric (rotation or mirroring) and/or a color-transformed version of an ARC task, i.e. a data augmentation process for tasks, which allows to create many ``new tasks'' for free and facilitates learning. Let us note that the same transform is applied to all training and test pairs of the task, which effectively creates a new task without leaking information about test samples or without breaking the fact that an underlying rule defines the task (albeit the said rule may differ from the original one without interfering with it during training, which is thus perfectly fine). 
If $D$ denotes the total number of untransformed demonstration pairs available in an optimization stage (pretraining or TTT) and $P$ is the average number of times each variant is presented for training, then the \emph{sample-presentation budget} is
\begin{equation}
    B = D \times V \times P = \text{Demos} \times \text{Variants} \times \text{Presentations}
    \label{eq:budget}
\end{equation}
which is the total number of training pairs encountered by the model during that stage. For TTT on a single evaluation task, $D$ is just the number of demonstration pairs in that task. For pretraining on \arcone{} training demonstrations, $D=1302$. With $D$ being fixed, one pair $(V,B)$ defines a budget configuration and $P$ adjusts implementation-specific epoch counts so that the budget configuration is matched in VARC's and TRM's respective codebases (epochs equal $P$ for VARC but must be computed as $P \times V$ for TRM). 

\subsubsection{Reference values from VARC and TRM.}
The budget configurations we will use are derived from the released methods. When pretraining on training tasks, VARC does not use variants, hence $V=1$, but uses RE-ARC extra data, making $D=400,000$, and uses $P=100$, hence $B=40M$. At test time, for a given task, VARC uses 50 augmentations in addition to the identity task, thus $V=51$, and with $P=100$ again, the sample-presentation budget is $B=5.1k \times D$, with $D$ typically equal to 3. With our notations, during pretraining on the training tasks, TRM uses 1000 variants, hence $V=1001$, $P=100$, and $D=1302$ since no extra data is used, giving $B=130M$, which is more than three times the training budget of VARC. At test time, for a given task, casting TRM in an inductive setting gives $V=1001$, $P=100$, and, for $D$ demos available (typically 3), gives $B=100.1k \times D$, which is 20 times more TTT budget than VARC. Hence, in summary, the $(V,B)$ reference configurations (with $D=3$ for TTT) are

\begin{equation}
  \begin{aligned}
  \mathrm{VARC\ pretrain}:~&(1,\,40\,\mathrm{M})
  &\qquad \mathrm{VARC\ TTT}:~&(51,\,15.3\,\mathrm{k}) \\
  \mathrm{TRM\ pretrain}:~&(1001,\,130\,\mathrm{M})
  &\qquad \mathrm{TRM\ TTT}:~&(1001,\,300.3\,\mathrm{k})
  \end{aligned}
  \label{eq:defaults}
\end{equation}

\subsection{Controlled Budget Grid}
\label{sec:configs}

Given the default VARC and TRM pretraining and TTT budgets (\cref{eq:defaults}), we considered all relevant combinations to measure budget effects on performance.

\subsubsection{Pretraining configurations.}
Our controlled pretraining always uses only the base ARC training demonstration pairs, so $D=1302$. We cross the two reference sample-presentation budgets with the two reference task-variant counts:
\begin{equation}
    \mathcal{C}_{\mathrm{pre}}
    = \{\text{none},
    (51,40\,\mathrm{M}), (1001,40\,\mathrm{M}),
    (51,130\,\mathrm{M}), (1001,130\,\mathrm{M})\},
\end{equation}
where each pair is a budget configuration $(V,B)$, and we also test no pretraining (``None''). Holding $B$ fixed while increasing $V$ necessarily reduces $P$: the model encounters more distinct task variants, but each variant is presented fewer times.

\subsubsection{TTT configurations.}
For TTT on an evaluation task, the number of demonstrations $D$ varies by task. With a typical value of $D=3$ (and adapting proportionally for other values), we consider
\begin{equation}
  \mathcal{C}_{\mathrm{TTT}}
  = \{
  (51,15.3\,\mathrm{k}), (1001,15.3\,\mathrm{k}),
  (51,300.3\,\mathrm{k}), (1001,300.3\,\mathrm{k})\},
\end{equation}
that is, we cross $V\in\{51,1001\}$ with a low budget of approximately 5.1k presentations per pair and a high budget of approximately 100.1k presentations per pair. Again $P$ is adapted so that the prescribed budget is respected. 

Crossing the five pretraining configurations with the four TTT configurations gives 20 budget configurations per model. VARC and TRM retain their own representations, optimizers, and decoding procedures; our experiment is designed to measure how pretraining budget and TTT budget affect each model family.

\subsection{Experimental Protocol}

\begin{figure}[H]
  \centering
  \includegraphics[width=\textwidth]{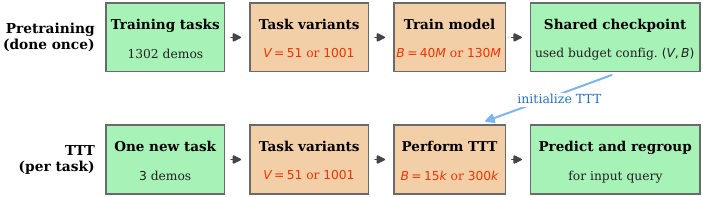}
  \caption{\textbf{Pretraining and TTT pipelines and budgets.} Pretraining on ARC tasks incurs a choice of task variants $V$ and of sample-presentation budget $B$, then it produces one pretrained model. Then, for each new evaluation task, a potentially different choice of variants $V$ can be made, as well as budget $B$ during the TTT process. The pretrained model is adapted for the task at hand, predicts the query, inverts the transformations, and aggregates predictions. Green boxes are fixed, orange boxes contain parameters that we vary, based on original VARC and TRM values.}
  \label{fig:controlled-protocol}
\end{figure}

\Cref{fig:controlled-protocol} summarizes the standardized part of the overall protocol. Offline pretraining is done once and produces a shared checkpoint. Independent per-task TTT initializes from that checkpoint, adapts to one new evaluation task at a time, produces ranked candidates, and then resets before the next task.

\subsubsection{Models.}
As a reminder, we experiment with two model families: VARC~\cite{hu2025arcvisionproblem}, which has 18M parameters and is inspired by Vision Transformers, but which we train without external RE-ARC data~\cite{hodel2023rearc}, contrary to the original paper; and TRM~\cite{jolicoeurmartineau2025morerecursivereasoningtiny}, which has 7M parameters and has two nested recursive loops that allow it to learn long sequences of operations at a modest parameter cost, but which we train in an inductive way, contrary to the original paper.

\subsubsection{Dataset and metrics.}
All experiments are assessed on \arcone{} evaluation tasks. We score the 400 tasks with task-level \passone{}, \passtwo{}, and oracle pass metrics explained in~\cref{sec:background}. We report \passtwo{} in the main text as it is the regular ARC metric used, and provide the others in~\cref{sec:complete-matrices}.

\subsubsection{TTT variants, aggregation and voting.}
For both models, after TTT on the task demonstrations, feedforward inference is run on that task's query inputs. The prediction of a task variant (e.g. a task that underwent a rotation) is then transformed back to the original task setup (e.g. rotated back), giving $V$ predictions per actual task. The frequency of each unique prediction is computed to rank predictions and calculate \passone{} and \passtwo{}.

\section{Results and Discussion}

\subsection{Results}
\label{sec:results}

\subsubsection{Performance per pretraining and TTT budget configurations.}

\begin{figure}[!htbp]
  \centering
  \includegraphics[width=\textwidth]{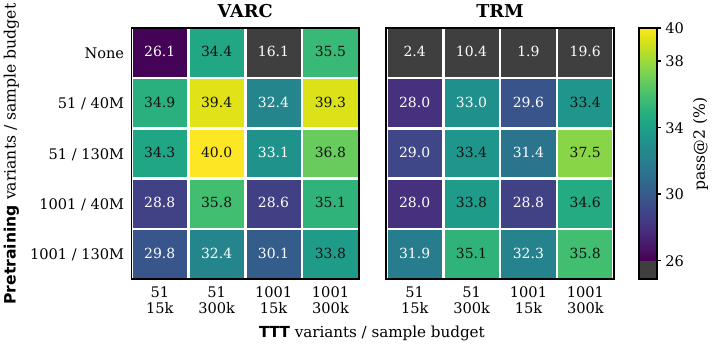}
  \caption{\textbf{Performance under fixed pretraining and TTT budgets.} We report \passtwo{} results for VARC and TRM. Rows vary the pretraining task-variant count $V$ and sample-presentation budget $B$; columns vary them for TTT. Both tables use a shared linear 26--40\% scale, with values below 26\% shown in dark grey.}
  \label{fig:protocol-heatmaps}
\end{figure}

\Cref{fig:protocol-heatmaps} provides the standard ARC \passtwo{} metric across all 40 results with the budget definitions from \cref{sec:configs}; \passone{} and oracle values are in~\cref{sec:complete-matrices}.

\subsection{Discussion}

\subsubsection{TTT sample-presentation budget is consistently important.}

Keeping the model, pretraining budget, and TTT task-variant count fixed, moving from the low (15.3 k) to the high (300.3 k) TTT sample-presentation budget improves \passtwo{} in all comparisons. Without pretraining, for VARC the gain is 8.26 points with 51 variants and 19.37 points with 1001 variants, and for TRM, these gains are 7.99 and 17.76 points. The low-budget 1001-variant runs from scratch indicate that using few sample presentations when having many variants might not give enough learning signal for the models, and the high-budget runs indicate that decent results can already be achieved without any pretraining.

\begin{figure}[!t]
  \centering
  \includegraphics[width=\textwidth]{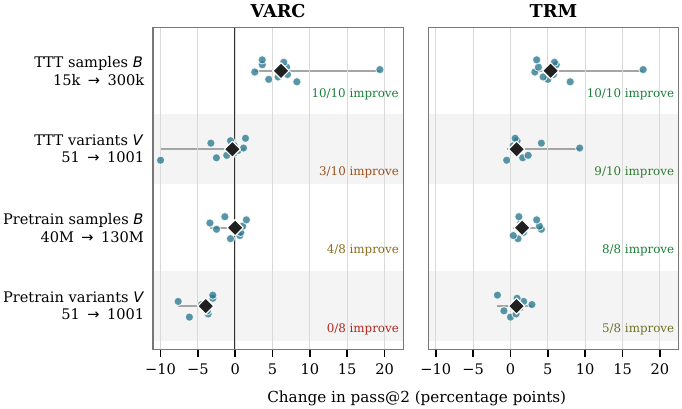}
  \caption{\textbf{Performance changes under budget increases.} We show the changes in \passtwo{} across our budget configurations when one budget is increased, indicated on the left. Diamonds show medians. Positive changes favor the budget-intensive setting.}
  \label{fig:protocol-factor-effects}
\end{figure}

Across all pretraining settings and both TTT task-variant counts, the median low-to-high gain is 6.13 points for VARC and 5.38 points for TRM. \Cref{fig:protocol-factor-effects} places this result beside the three other controlled changes in the grid. Each point changes only the reported factor while keeping the other factors fixed. The TTT sample-presentation budget is the only uniformly beneficial factor for both models; the effects of TTT task-variant count, pretraining sample presentations, and pretraining task-variant count depend on the model and surrounding budgets.

\subsubsection{TTT task-variant count benefits mostly TRM.}

Across both TTT sample-presentation budgets, moving from 51 to 1001 variants improves VARC \passtwo{} in three of ten matched comparisons and TRM in nine of ten; the median changes are $-0.37$ and $+0.82$ points, respectively, as shown in \Cref{fig:protocol-factor-effects}. At high TTT budget, 1001 variants improve TRM in every pretraining row, by 0.37 to 9.26 points, and improve oracle by 5.00 to 14.13 points (see Appendix), meaning that the correct output grid is much more often proposed in the candidate prediction list. VARC behaves differently. At low budget, 1001 variants reduce \passtwo{} in four of five rows. At high budget, 51 variants perform better in three rows and 1001 variants in two, with differences ranging from $-3.25$ to $+1.38$ points for the 1001-variant setting; oracle nevertheless improves in every row by 1.63 to 3.88 points. For VARC, a larger TTT task-variant count therefore broadens the candidate list more reliably than it improves the two-candidate score; for TRM, it also improves \passtwo{} when enough sample presentations are available.

\subsubsection{Pretraining helps especially TRM, but more is not always better.}

As shown in~\Cref{fig:protocol-heatmaps}, pretraining on ARC training tasks improves all 16 TRM cells over no pretraining. The best TRM \passtwo{} rises from 19.63\% without pretraining to 37.50\% after 51-variant/130M pretraining. VARC starts from a stronger no-pretraining baseline: pretraining improves 13 of its 16 cells. Its best \passtwo{} rises from 35.50\% to 40.00\% after 51-variant/130M pretraining.

We do not observe a ``more pretraining is better'' rule. Raising the pretraining sample-presentation budget from 40M to 130M improves VARC \passtwo{} in four of eight cases, and all eight for TRM. Increasing the pretraining task-variant count from 51 to 1001 is harmful for VARC in all eight comparisons and modestly improves five of eight comparisons for TRM. The appropriate pretraining budget is thus model-dependent and must be evaluated jointly with TTT.

\subsubsection{Note on compute time.}

Sample-presentation budget and measured compute time are related but distinct. Pretraining is, in practice, done once and can then be reused, whereas TTT is repeated for every evaluation task. Our complete 400-task logs for TTT on A100 GPUs show that low-budget TTT costs 12.46--18.27 GPU-hours for VARC and 9.78--13.66 GPU-hours for TRM; high-budget TTT costs 190.75--198.37 and 148.67--152.62 GPU-hours, respectively. Thus, the consistent performance gain from a larger TTT sample-presentation budget comes with an order-of-magnitude increase in evaluation cost. Detailed measurements in \cref{sec:compute} can guide budget selection under limited academic compute.

\subsubsection{Limitations.}

Our results cover only \arcone{} because both VARC and TRM showed strong results on it (54.5\% and 44.6\% respectively)~\cite{hu2025arcvisionproblem,jolicoeurmartineau2025morerecursivereasoningtiny}, so we were confident we had enough room to experiment with less favorable training conditions (no RE-ARC for VARC, no transductive setting for TRM) while still obtaining decent and valuable results. Besides, each result reports only one experimental run, hence small performance differences should be interpreted cautiously, and the main conclusions are on the larger and consistent changes across configurations. Moreover, because VARC and TRM differ substantially in parameter count, our results measure budget sensitivity within each family rather than providing a ranking between the two architectures. While variants exist or can be built for each model, we chose to use the reference architectures as main baselines. Furthermore, extending our study to other datasets, such as ARC-AGI-2~\cite{chollet2026arcagi2newchallengefrontier}, could be an interesting follow-up but, as this benchmark is notoriously more difficult, VARC and TRM only score 8.3\% and 7.8\% on it~\cite{hu2025arcvisionproblem,jolicoeurmartineau2025morerecursivereasoningtiny}, hence we refrained from burning a significant amount of academic GPU resources for arguably weak results overall. The same goes for PotARCin~\cite{beger2026potarcinmultidimensionalevaluationskill}, which challenges frontier models at generalization in various dimensions absent from ARC. We are confident that our observations should remain valid for some other ARC-like datasets, such as Compositional-ARC~\cite{mondorf2026compositionalarc} and Sort-of-ARC~\cite{assouel2022objectcentric}, which target compositionality in abstract visual reasoning, or ConceptARC~\cite{moskvichev2023the}, which evaluates conceptual understanding in the ARC domain, but we leave that verification for future works. Other visual reasoning problems, such as Bongards~\cite{bongard1970pattern,langenfeld2026bongardsboundaryperceptionreasoning}, are more distant and might require significant architecture changes, which we also leave for future work.

\section{Conclusion}
\label{sec:conclusion}

Our study shows that pretraining and test-time training (TTT) budgets largely influence ARC performance within the VARC and TRM models, which was not clear from their original releases, which relied, for VARC, on external training data and, for TRM, on a transductive setting. With a controlled ARC-only inductive setup, we observe that increasing the TTT sample-presentation budget (i.e. how many samples are seen during TTT) helps both models consistently; increasing the task-variant count (i.e. a heavier task augmentation) mainly helps TRM; and pretraining is especially beneficial to TRM while VARC is already capable with TTT alone. These findings do not rank the two architectures. As VARC and TRM triggered two separate lines of research on ARC, our results underline that ARC papers should better standardize their training budgets to understand whether performance gains might come from architectures alone.

\begin{credits}
\subsubsection{Acknowledgements}
Adrien Deli\`ege is a F.R.S.-FNRS post-doc. This research benefited from the computational resources of Lucia, the Tier-1 supercomputer of the Walloon Region, funded under grant no.~1910247.

\subsubsection{Disclosure of Interests}
The authors declare no competing interests.
\end{credits}

\appendix

\section{Complete Performance Tables}
\label{sec:complete-matrices}

In \cref{tab:complete-matrices}, we provide the complete \passone{} / \passtwo{} / oracle performance over the 400 evaluation tasks.

\begin{table}[H]
  \caption{Complete \arcone{} tables. Each cell is \passone{} / \passtwo{} / oracle in percentage over 400 evaluation tasks. Rows are pretraining task-variant count/sample-presentation budget; columns are the corresponding TTT budgets. All pretrained rows use ARC training demonstrations without RE-ARC.}
  \label{tab:complete-matrices}
  \centering
  \scriptsize
  {\setlength{\tabcolsep}{3.2pt}\renewcommand{\arraystretch}{1.08}
  \begin{tabular}{@{}lcccc@{}}
    \toprule
    \multicolumn{5}{c}{\textbf{(a) VARC}} \\
    Variants/Samples & TTT 51/15.3k & TTT 51/300.3k & TTT 1001/15.3k & TTT 1001/300.3k \\
    \midrule
    No pretraining & 23.75/26.12/34.75 & 30.75/34.38/43.13 & 14.50/16.13/23.13 & 30.63/35.50/46.25 \\
    Pretrain 51/40M & 31.75/34.88/43.00 & 35.88/39.37/50.00 & 29.13/32.37/47.63 & 33.87/39.25/52.62 \\
    Pretrain 51/130M & 31.50/34.25/42.25 & 35.00/40.00/47.75 & 29.38/33.12/44.75 & 33.00/36.75/49.38 \\
    Pretrain 1001/40M & 26.50/28.75/39.25 & 33.12/35.75/46.25 & 24.87/28.63/44.00 & 30.63/35.13/48.88 \\
    Pretrain 1001/130M & 26.62/29.75/38.25 & 28.88/32.37/39.75 & 26.12/30.13/41.50 & 30.25/33.75/43.63 \\
    \bottomrule
  \end{tabular}

  \vspace{0.4em}

  \begin{tabular}{@{}lcccc@{}}
    \toprule
    \multicolumn{5}{c}{\textbf{(b) TRM}} \\
    Variants/Samples & TTT 51/15.3k & TTT 51/300.3k & TTT 1001/15.3k & TTT 1001/300.3k \\
    \midrule
    No pretraining & 1.87/2.38/4.25 & 8.75/10.37/15.12 & 1.38/1.87/3.38 & 18.25/19.63/29.25 \\
    Pretrain 51/40M & 25.12/28.00/34.88 & 29.13/33.00/39.12 & 27.12/29.63/40.88 & 30.50/33.37/50.38 \\
    Pretrain 51/130M & 24.62/29.00/34.38 & 30.13/33.37/39.75 & 28.75/31.37/42.63 & 35.38/37.50/50.00 \\
    Pretrain 1001/40M & 23.75/28.00/36.00 & 30.13/33.75/40.00 & 26.25/28.75/42.75 & 31.87/34.63/45.00 \\
    Pretrain 1001/130M & 27.25/31.87/38.37 & 32.00/35.13/42.25 & 29.75/32.25/44.50 & 32.25/35.75/48.25 \\
    \bottomrule
  \end{tabular}}
\end{table}

A technical detail is that, for VARC, we retain its released ten-view inference procedure, so each of its $V$ task variants yields ten predictions before inversion and voting ($10V$ candidates in total), whereas TRM yields one prediction per task variant ($V$ candidates). Candidate-generation budgets are therefore model-specific, thus oracle values should be interpreted primarily within each model family rather than as a direct comparison between families. Still, preliminary experiments with 1-view prediction per variant for VARC and with 10-view predictions for TRM did not seem to yield significantly different results, hence we did not further explore that direction.

\section{Compute Cost}
\label{sec:compute}

We distinguish the sample-presentation budget from measured compute cost. Pretraining is done once and can be amortized across tasks and later experiments, whereas TTT is done again for every evaluation task. Both costs matter, but they are not interchangeable: $B$ measures the number of training pairs encountered, while GPU-hours also depend on the model, implementation, hardware, and inference procedure.

\Cref{tab:pretrain-runtime} reports the pretraining runtimes on nodes of 8 $\times$ A100 GPUs. These values should be interpreted together with model size and implementation efficiency, not as a hardware-normalized architecture comparison.

\begin{table*}[tb]
  \caption{ARC pretraining compute on 8-A100 GPUs. GPU-hours are wall time multiplied by eight. The \passtwo{} range covers the four downstream TTT cells.}
  \label{tab:pretrain-runtime}
  \centering
  \small
  \begin{tabular*}{\textwidth}{@{\extracolsep{\fill}}lrrrrr@{}}
    \toprule
    Model & $V$ & $B$ & Wall-h & GPU-h & \passtwo{} range \\
    \midrule
    VARC-ViT & 51 & 40M & 10.81 & 86.5 & 32.37--39.37 \\
    VARC-ViT & 51 & 130M & 35.08 & 280.6 & 33.12--40.00 \\
    VARC-ViT & 1001 & 40M & 12.12 & 96.9 & 28.63--35.75 \\
    VARC-ViT & 1001 & 130M & 38.61 & 308.9 & 29.75--33.75 \\
    TRM & 51 & 40M & 5.3 & 42.4 & 28.00--33.37 \\
    TRM & 51 & 130M & 16.77 & 134.2 & 29.00--37.50 \\
    TRM & 1001 & 40M & 7.89 & 63.1 & 28.00--34.63 \\
    TRM & 1001 & 130M & 21.1 & 168.8 & 31.87--35.75 \\
    \bottomrule
  \end{tabular*}
\end{table*}

For \cref{tab:runtime}, each runtime is the mean across complete 400-task TTT runs with the same budget configuration but different pretraining checkpoints. GPU-hours sum their elapsed times, and time per task divides that total by 400. Each TTT used one A100 GPU. Scratch \passtwo{} is shown as a common performance reference, but is not averaged across the runtime runs.

\begin{table*}[tb]
  \caption{TTT runtime and scratch performance for all TTT configurations. Runtime values are means across complete 400-task TTT runs on A100 GPUs, with one GPU per running task. Scratch \passtwo{} is reported as a common performance reference.}
  \label{tab:runtime}
  \centering
  \small
  \begin{tabular*}{\textwidth}{@{\extracolsep{\fill}}lrrrrr@{}}
    \toprule
    Model & $V$ & $B$ & Time/task & GPU-h & Scratch \passtwo{} \\
    \midrule
    VARC & 51 & 15k & 112.1s & 12.46 & 26.12 \\
    VARC & 51 & 300k & 1716.8s & 190.75 & 34.38 \\
    VARC & 1001 & 15k & 164.4s & 18.27 & 16.13 \\
    VARC & 1001 & 300k & 1785.3s & 198.37 & 35.50 \\
    TRM & 51 & 15k & 88.0s & 9.78 & 2.38 \\
    TRM & 51 & 300k & 1338.1s & 148.67 & 10.37 \\
    TRM & 1001 & 15k & 123.0s & 13.66 & 1.87 \\
    TRM & 1001 & 300k & 1373.6s & 152.62 & 19.63 \\
    \bottomrule
  \end{tabular*}
\end{table*}

Across the two complete $5 \times 4$ performance tables, the successful runs used approximately 4,904 A100 GPU-hours. Pretraining the four checkpoints for each model required 1,181.4 GPU-hours in total: 772.9 for VARC and 408.5 for TRM. Each of the four TTT budget configurations was then evaluated under five initialization conditions, namely from scratch and from each of the four pretrained checkpoints, yielding 40 complete 400-task TTT runs across both models. TTT required approximately 3,722.9 GPU-hours: 2,099.3 for VARC and 1,623.7 for TRM. The complete tables therefore cost approximately 2,872.2 GPU-hours for VARC and 2,032.2 GPU-hours for TRM, equivalent to 204 single-GPU days overall. TTT represented about 76\% of this measured compute. Moreover, the high sample-presentation TTT configurations alone accounted for approximately 3,452 GPU-hours, or 93\% of the total TTT cost, illustrating the enormous cost of increasing the per-task training budget in practice. These values cover the successful runs producing the main results and exclude failed jobs, preliminary experiments, diagnostic analyses, and development runs.

\bibliographystyle{splncs04}
\bibliography{main}
\end{document}